\documentclass[Journal]{IEEEtran}
\IEEEoverridecommandlockouts
\usepackage{cite}
\usepackage{amsmath,amssymb,amsfonts}
\usepackage{algorithmic}
\usepackage{graphicx}
\usepackage{textcomp}
\usepackage{xcolor}
\usepackage{algorithm}
\usepackage{algorithmic}
\usepackage{subfigure}
\def\BibTeX{{\rm B\kern-.05em{\sc i\kern-.025em b}\kern-.08em
    T\kern-.1667em\lower.7ex\hbox{E}\kern-.125emX}}
\begin{document}

\title{An In-Vehicle Keyword Spotting System with Multi-Source Fusion for Vehicle Applications\\
}

\author{
\IEEEauthorblockN{Yue Tan\IEEEauthorrefmark{1}, Kan Zheng\IEEEauthorrefmark{1}, Lei Lei\IEEEauthorrefmark{2}}

\IEEEauthorblockA{\IEEEauthorrefmark{1}Intelligent Computing and Communication (IC$^2$) Lab, \\
Key Laboratory of Universal Wireless Communications, Ministry of Education,\\
Beijing University of Posts and Telecommunications, Beijing, China, }\\
\IEEEauthorblockA{\IEEEauthorrefmark{2}James Cook University, Australia}

tanyue@bupt.edu.cn}
\maketitle
\begin{abstract}
In order to maximize detection precision rate as well as the recall rate, this paper proposes an in-vehicle multi-source fusion scheme in Keyword Spotting (KWS) System  for vehicle applications. Vehicle information, as a new source for the original system, is collected by an in-vehicle data acquisition platform while the user is driving. A Deep Neural Network (DNN) is trained to extract acoustic features and make a speech classification. Based on the posterior probabilities obtained from DNN, the vehicle information including the speed and direction of vehicle is applied to choose the suitable parameter from a pair of sensitivity values for the KWS system. The experimental results show that the KWS system with the proposed multi-source fusion scheme can achieve better performances in term of precision rate, recall rate, and mean square error compared to the system without it.
\end{abstract}


\section{Introduction}
Keyword Spotting (KWS) System, also known as wake-word detection, refers to the task of detecting specified keyword from a continuous stream of audio provided by the users \cite{b1}. Keyword Spotting has been an active research area in speech recognition for decades, and widely used in numerous applications. Typical applications exist in environments with interference from background audio, reverberation distortion, and the sounds generated by the device in which the KWS is embedded. Nowadays, conversational human-technology interfaces become increasingly popular in a large number of applications. There are already millions of devices with embedded KWS systems. With the advances in deep learning and increase in the amount of available data, traditional approaches for KWS has been replaced by deep-learning-based approaches due to their superior performance.

As KWS is the most common way for human to interact with machine in vehicle and determines different states of the device or software system, the performance of KWS system is very crucial for vehicle applications \cite{b2}. Although there are different kinds of interferences, it is expected that the output of the KWS system has a high detection precision rate and recall rate to meet the users’ requirements.

The Hidden Markov Model (HMM) have been widely used in traditional KWS systems \cite{b3}. When the likelihood ratio of keyword model to background model exceeds the threshold, the system is triggered \cite{b4}. Also, Gaussian Mixture Models (GMM) are used to establish acoustic models. In recent years, deep neural network (DNN) is widely used in acoustic modeling and better performs than GMM \cite{b5}. Later, the systems based on a Convolutional Neural Network (CNN) or Deep Neural Network (DNN) instead of HMM, become popular since the amount of available data has been rapidly increased \cite{b6}. Although the above methods based on neural network have achieved good performance, there are still a few defects, e.g., the audio input has a strong dependency in time or frequency domains,  the system needs broadband filters to model the context over the entire frame in terms of CNN \cite{b7}. Other related works explore discriminative models for keyword spotting based on large-margin formulation or recurrent neural networks \cite{b12}\cite{b13}. However, the large-margin formulation based methods require processing of the entire utterance to find the optimal keyword region. This increases detection latency. Although some complicated structure of neural network are proposed to deal with them, the huge computation complexity becomes a huge challenge for real-time applications \cite{b3}\cite{b8}.

Most KWS systems use a single source of information as their input, e.g., the audio. However, in many applications, e.g., vehicle applications, the environmental information and sensor data are always available and easy to obtain, which may provide additional clues for KWS \cite{b16}. In order to improve the performance of KWS system for a keywords detecting task, these data can be used as a complement to the audio information. This multi-source fusion is accomplished by adjusting the parameters of DNN-based KWS system using the additional information.

In vehicles, the KWS system is used for a wide range of vehicle applications, which are crucial to the intelligent vehicular safety systems. Therefore, in this paper, we propose an in-vehicle KWS system with the multi-source fusion for vehicle applications. In this system, raw vehicle information is collected by an in-vehicle data acquisition platform \cite{b9}\cite{b15}. The single-source system uses DNN to detect keywords and outputs the confidence score. The processed vehicle information is used as an additional source of the system and fused with DNN-based single-source KWS system. Thus, it provides the possibility for KWS system to make use of the dynamic vehicle information such as speed and direction when detecting the keywords spoken by driver. During this process, a better sensitivity value for the system is chosen by multi-source fusion algorithm. An appropriate sensitivity value can help KWS system achieve a higher accuracy when detecting keywords.

The rest of this paper is organized as follows: Section II describes the overall architecture of our system. Section III presents the implementation of the system in detail. In Section IV, we give the experimental results and analysis. Finally, conclusions are provided in Section V.
\section{System Overview}
To improve the performance of in-vehicle KWS system, the real-time vehicle information is obtained and fed back to the KWS system \cite{b2}. Then, the system chooses suitable parameters, e.g., sensitivity value, and outputs the classification result. Fig.1 depicts the structure of an in-vehicle KWS system with multi-source fusion for vehicle applications. Five components are included as follows:

\begin{figure}
\begin{center}
\includegraphics[width=3in]{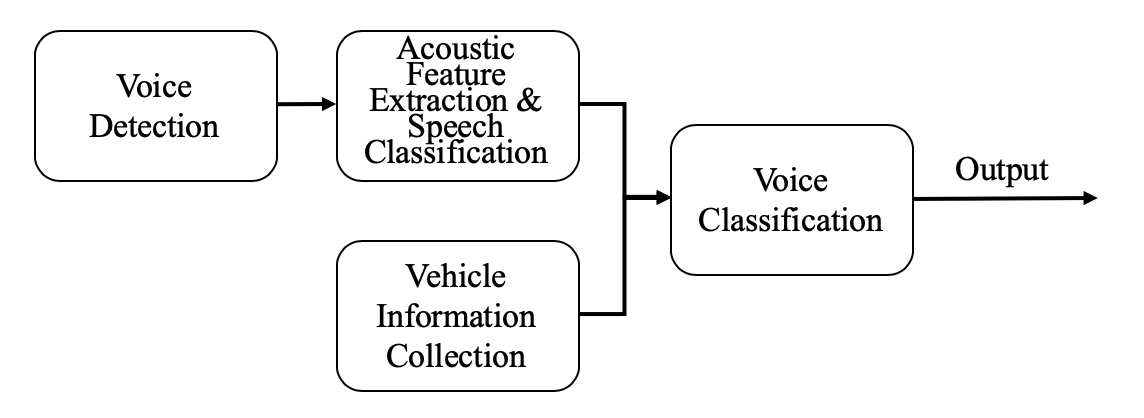}
\caption{Structure of in-vehicle KWS system with multi-source fusion} \label{fig1}
\end{center}
\end{figure}

\subsection{Voice Detection}

The initial step of the whole system is voice detection and pretreatment. The voice input is a continuous stream of audio from people. First, microphone converts the voice input into an instantaneous waveform sampling flow which is then converted to a sequence of frames by spectrum analyzer. After that, a certain number of frames are sent to the next step for feature extraction as a unit \cite{b10}.

During voice detection and preprocessing, the frame length $t_0$, and the number of frames $n$ are two variables involved. For example, when the vehicle is in complex traffic situations, $t_0$ should be shorter and $n$ should be larger.

The source of sound comes from microphone of mobile terminal in real-time working state. When the microphone is opening, it continuously extracts information from voice streams.
\subsection{Acoustic Feature Extraction \& Speech Classification}

The frames obtained in Step A are fed into the acoustic model, and acoustic features are extracted over the frames \cite{b14}. Then, the acoustic features are stacked as a larger vector, which is fed as input to the DNN. The structure of DNN is shown in Fig.2. DNN is trained to predict posterior probabilities for each output label from the stacked features. Output labels correspond to the keywords or some similar words. Finally, a simple posterior handling module combines the posteriors into a confidence score used for detection \cite{b1}.

\begin{figure}
\begin{center}
\includegraphics[width=3in]{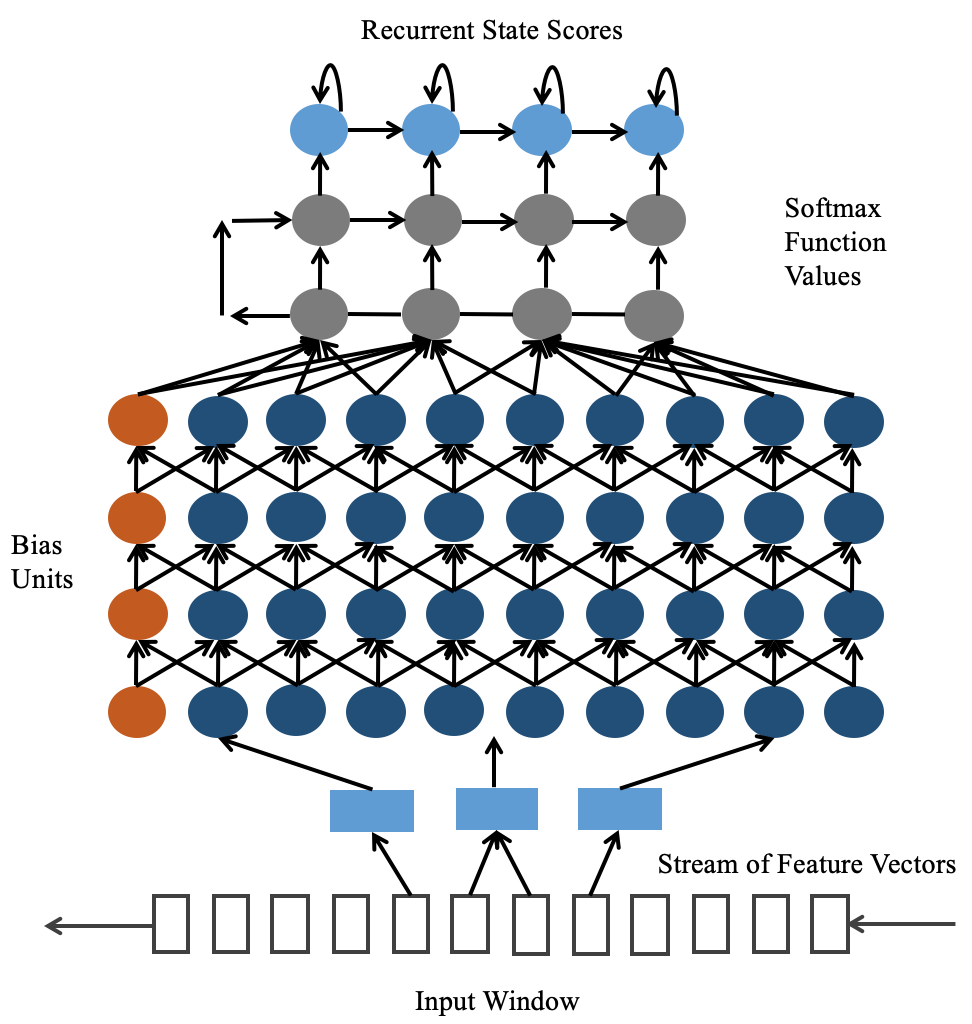}
\caption{ The structure of DNN} \label{fig2}
\end{center}
\end{figure}

\subsection{Vehicle Information Collection}

In vehicles, raw vehicle information is collected by an in-vehicle data acquisition platform. Vehicle information, as a new source for the single-source system, is collected while the user is driving. Raw vehicle information includes the longitude and latitude of the vehicle and is used to compute the speed and direction of vehicle. In order to improve the predictive precision rate and recall rate, the processed vehicle information is applied to choose suitable parameter, e.g., sensitivity value, for the original KWS system.

\subsection{Voice Classification}

After combining with vehicle information, the system can choose the better one from a pair of sensitivity values. According to the speech classification result of DNN, voice classification result can be obtained. If the voice classification result suggests that the voice input is more likely to be the keywords being detected, related operating hardware and software can be activated. On the contrary, if the result indicates the voice input is related to other voice classification, nothing responses to the input.

\section{Implementation of KWS System with Multi-Source Fusion}
In this section, the implementation of KWS system with multi-source fusion is presented in detail. The DNN-based single-source algorithm is conducted on a single-source voice detection system which uses 13-dimensional features and the features' deltas as input to a trained model. The model is a 30-component diagonal covariance GMM, which generates posteriors to determine whether it is speech or not at every frame. After that, a number of frame posteriors are sent to a state machine. If most of them exceed a threshold, those frames are identified as speech regions. For speech regions, acoustic features are generated based on 40-dimensional log filterbank energies. Features are extracted every 10 $ms$ over a window of 30 $ms$. Since each frame adds 10 $ms$ of latency to KWS system, the input window is not symmetric. For the single-source DNN-based KWS system, we use 10 future frames and 30 frames in the past.

The DNN model used for KWS is a feed-forward fully connected neural network which includes $k$ hidden layers and $n$ hidden nodes every layer. Each node computes a non-linear function of the weighted sum of the previous layer's output. For the hidden layers, Rectified Linear Unit (ReLU) function is used because it performs better than logistic function. The last layer has a softmax which outputs an estimate of the posterior of each output label. The size of the network is also decided by the number of output labels.

Suppose $p_{ij}$ is the posterior for the $i^{th}$ label and the $j^{th}$ frame $x_j$, and $n$ is the number of frames. In $p_{ij}$, $i$ takes values between $0, 1, \cdots, n-1$, where $0$ refers to the label which cannot be related to any part of the keywords. $\theta$ is used to represent the weights and biases of the deep neural network and are computed by minimizing the cost function over training data $(x_j, i_j)$. For the DNN-based KWS, the labels represent the entire keywords or part of them.

DNN has produced label posteriors which are based on the frames. Those posteriors are related to confidence scores. If the confidence score exceeds the threshold, the system can be activated.

First, we need to smooth the posteriors over a time window of size $w_{s}$. The posterior is denoted as $p_{ij}$, and smoothed posterior is denoted as $p^{'}_{ij}$. The smoothing result is given by (1),

\begin{equation}
p^{'}_{ij}=\frac{1}{j-h_{s}+1}\sum_{k=h_{s}}^{j}p_{ik}
\label{eq1}
\end{equation}

\noindent where $h_{s}=\max\{1,j-w_{s}+1\}$, which means that $h_s$ is the index of the first frame within the smoothing window \cite{b17}.

Then, the confidence score of $j^{th}$ frame is computed over the sliding window of size $w_{max}$.

\begin{equation}
score=\sqrt[n-1]{\prod_{i=1}^{n-1}\max_{h_{max}\leq k\leq j}{p^{'}_{ik}}}
\label{eq2}
\end{equation}

\noindent where $p^{'}_{ik}$ is the smoothed posterior in (1) and $h_{max}=\max\{1,j-w_{max}+1\}$ is the index of the first frame within the sliding window. Generally, the value of $w_s$ and $w_{max}$ are set according to the set. However, the performance is not very sensitive to $w_s$ and $w_{max}$. Therefore, the multi-source fusion algorithm is proposed to help the system choose better sensitivity value \cite{b11}. How the vehicle information is processed to execute the algorithm is explained as follow.

When the vehicle is moving, a set of real-time information is collected for analysis. The data includes latitude, longitude and altitude of the vehicle and the timestamp of the record time. Vehicle information can be calculated on the basis of this raw data.

The workflow of multi-source fusion is depicted in Algorithm 1. The input consists of vehicle information set $D$, a continuous stream of audio $S$, double sensitivity values $sen_1$, $sen_2$ and the thresholds of the variation of speed and direction $s_{thd}$, $d_{thd}$. As mentioned in Section II, firstly, the stream of audio $S$ is converted to a sequence of frames $\varphi(S)$. Then, a certain number of frames are fed as input to DNN. DNN extracts features over these frames, which are denoted as $(f_1,f_2, \cdots f_m)=F(\varphi(S))$. Meanwhile, vehicle information set $D$ is preprocessed to provide additional source for the system, which is denoted as AS: $(s_1,d_1 ),(s_2,d_2 ), \cdots (s_m,d_m )$. For each $(s_m,d_m )$, we calculate $\Delta$$s$ and $\Delta$$d$ and compare to $s_{thd}$ and $d_{thd}$, respectively. Finally, the sensitivity value of KWS system is determined by $\Delta$$s$ and $\Delta$$d$. The result of KWS is generated with one of the sensitivity value.

\begin{algorithm}[t]
	\caption{Multi-source fusion algorithm}
    \hspace*{0.02in} {\bf Input:} 

    \hspace*{0.06in}A continuous stream of audio, $S$

    \hspace*{0.06in}Double sensitivity, $sen_1$, $sen_2$  ($sen_1$$>$$sen_2$)

    \hspace*{0.06in}Threshold of the variation of speed and direction, $s_{thd}$, $d_{thd}$

    \hspace*{0.06in}Vehicle information set, $D$: ($Lng_1$, $Lat_1$), ($Lng_2$, $Lat_2$), $\cdots$ ($Lng_m$, $Lat_m$)

    \hspace*{0.02in} {\bf Output:} 

	\begin{algorithmic}[1]
		\STATE Convert $S$ to a sequence of frames $\varphi(S)$
        \STATE Extract features from $\varphi(S)$, $(f_1,f_2, \cdots f_m)=F(\varphi(S))$
        \STATE Preprocess $D$ to get speed and direction as an additional source $AS$: $(s_1,d_1 ),(s_2,d_2 ), \cdots (s_m,d_m )$
		\FOR{$s_i, d_i$ in $AS$}
		\IF{$\Delta$$s$$>$$s_{thd}$ and $\Delta$$d$$>$$d_{thd}$}
		\STATE generate Result for $S$ with $sen_2$
        \ELSE
		\STATE generate Result for $S$ with $sen_1$
		\ENDIF
		\ENDFOR
	\end{algorithmic}
\end{algorithm}

In complex traffic situations, drivers are more frequently involved in traffic accidents, especially for inexperienced drivers. As for this situation, drivers are more likely to ask for help from the navigation system by using KWS system. As an additional source, the vehicle information is used for the selection of better sensitivity. Since the KWS with multi-source fusion dynamically tunes the sensitivity value, the system in vehicle improves the predictive recall rate and real-time rate.

It is assumed that the system has two kinds of states, normal state and sensitive state. When the system is in normal state, the recall rate is $p_1$, and while in sensitive state, the recall rate is $p_2$. It is obvious that $p_2$$>$$p_1$. The average time proportion of sensitive state is supposed to be $k$. It is presumed that the probability that the in-vehicle data acquisition platform outputs the correct information is $p_3$.

Based on this assumption, the recall rate of single-source KWS system is

\begin{equation}
recall_1=p_1
\label{eq3}
\end{equation}

The recall rate of multi-source KWS system is

\begin{equation}
\begin{aligned}
recall_2=&p_1 \cdot [(1-k) \cdot p_3+k \cdot (1-p_3 )]+ \\
&p_2 \cdot [(k \cdot p_3+(1-k)(1-p_3 ))]
\end{aligned}
\label{eq4}
\end{equation}

(\ref{eq4}) $-$ (\ref{eq3}) :

\begin{equation}
\begin{aligned}
recall_2-recall_1=&(p_2-p_1 )[(1-k)(1-p_3 )+ \\
&k \cdot p_3 ]>0
\end{aligned}
\label{eq5}
\end{equation}

It can be seen from (\ref{eq5}) that the recall rate of KWS system is improved after multi-source fusion scheme is used.

When the KWS system is in sensitive state, the length of each frame $t_0$ gets shorter, which contributes to the enhancement of real time rate. The state change of vehicle is precisely captured by the system. It is hard for a single-source system to do that because we cannot find an appropriate sensitivity value for all the states. Although DNN applied here has already reached a high accuracy, it is difficult for a single sensitivity value to satisfy all potential state.

\section{Experimental Results and Analysis}

\subsection{Experimental Environment}
The in-vehicle data acquisition platform runs on any iOS-based smart portable devices and collects sensor information. A few roads near Beijing University of Posts and Telecommunications are chosen as test roads to simulate the trajectories of a normal driver. These trajectories include making a turn, turning around, entering a roundabout, etc. The experiments are conducted in the mobile terminal. To avoid noise distractions, the test is carried out in a quiet environment.

\subsection{Test Set}
Experiments are performed on a test set which combines real voice including the keywords as positive examples and phrase including parts of the keywords or some similar words as negative examples \cite{b1}. The test set is generated by: (1) generating noise by simulating the vehicle environment; (2) recording the keywords; (3) adjusting the characteristic of recording to expand the test set.

First, white Gaussian noise with different power is generated to simulate the vehicle environment. Then, the experimental content is recorded, which consists of the keywords, words pronounced similar to keywords, everyday words and common phrase for driving. There are in total four possible speakers with different pronunciation characteristics. The gender ratio of the speakers is $1:1$.

On the basis of raw audio files, a professional audio processing software\emph{cool edit pro} is employed to adjust the tone and speed of each fragment. In this way, multiple samples are obtained. In this way, the test set of both positive and negative examples is expanded. Finally, the audio test set consists of 50 positive examples and 50 negative examples. The keywords is “\emph{HEY, ATOM}”, and related negative examples include “\emph{HEY, ATO}”, “\emph{HEY, TOM}”, “\emph{HELLO, ATOM}”, etc.

The voice segments of positive and negative cases are mixed with Gaussian white noise. The range of signal-to-noise (SNR) ratio is 5-10 $dB$. A fragment of test set which combines human voice and white Gaussian noise is shown in Fig.3.

\begin{figure}
\begin{center}
\includegraphics[width=3in]{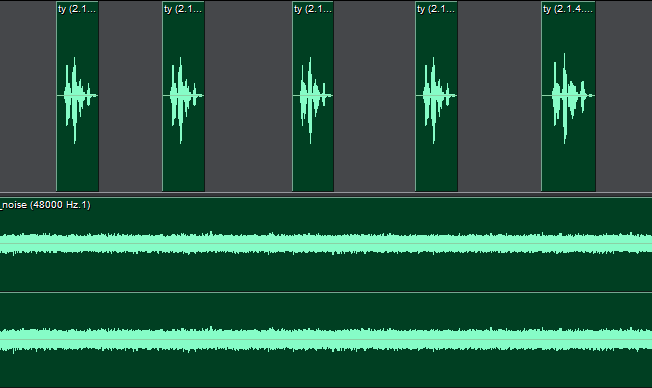}
\caption{A fragment of test set} \label{fig3}
\end{center}
\end{figure}

\subsection{Experimental Results}
The vehicle information are visualized in Fig.4, in which the horizontal axis presents the displacement from west to east and the vertical axis presents the displacement from north to south. Speed is presented by the scatters in different colors, and the unit of speed is $m/s$. When the vehicle is approaching the intersection, it always decreases the speed at first and then increases it.

\begin{figure}
  \centering
  \subfigure[]{
    \includegraphics[width=2.5in]{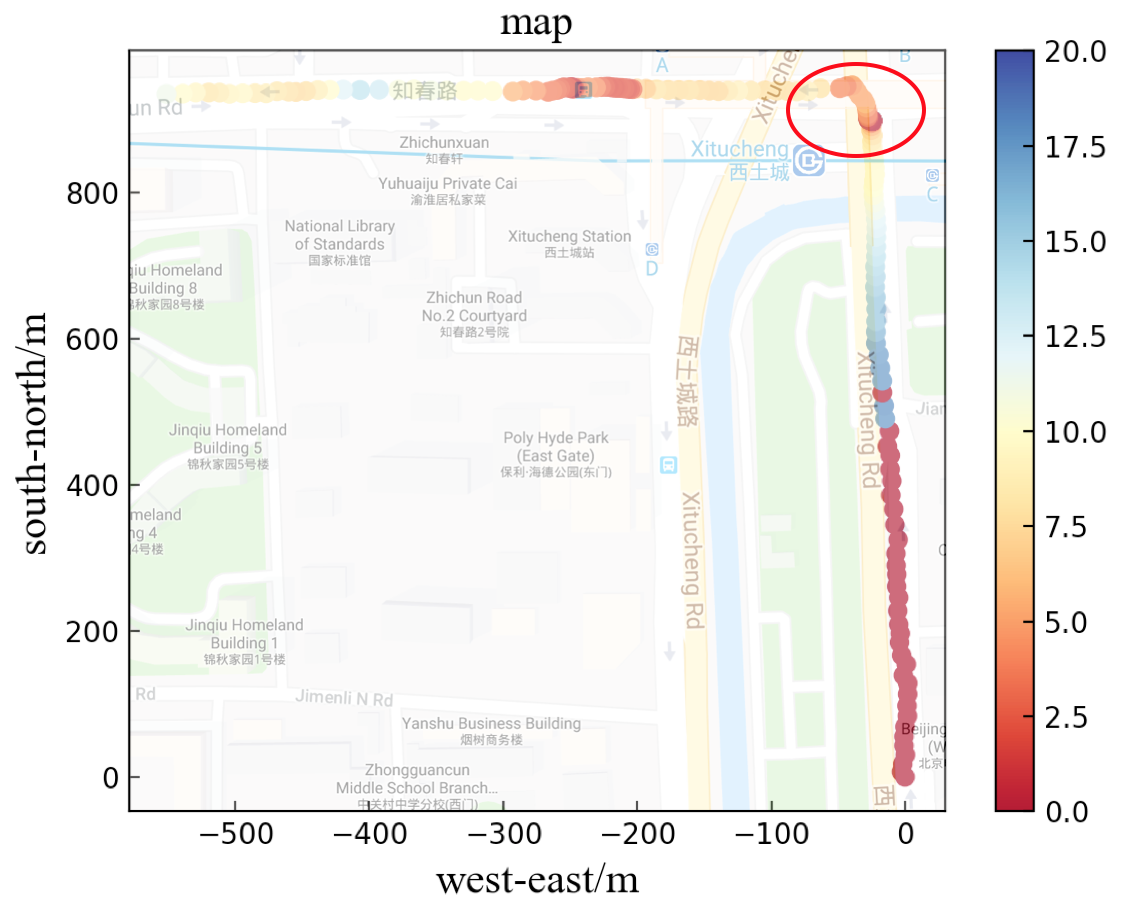}
  }

  \subfigure[]{
    \includegraphics[width=2.6in]{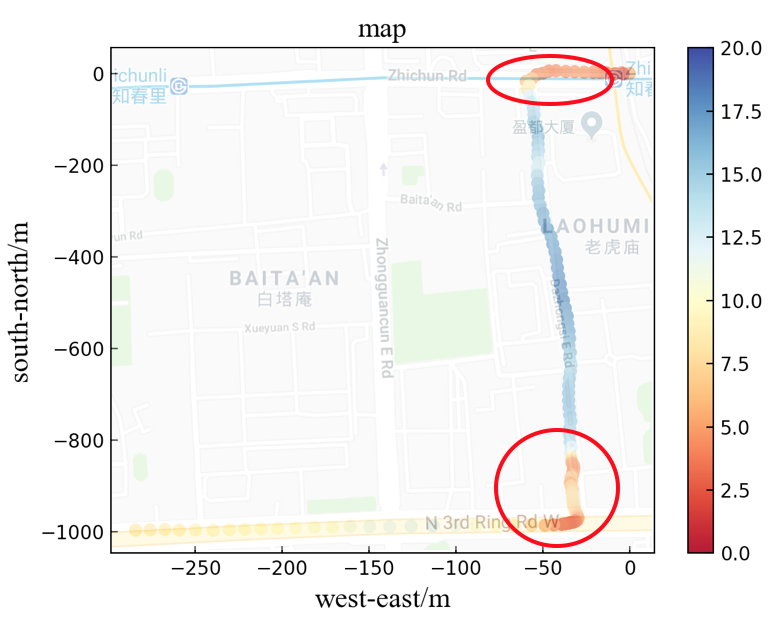}
  }
  \caption{Illustration of the variation of speed and position during intersection turning}
\end{figure}

Similarly, in Fig.5, when the vehicle is turning around, it slows down to nearly zero speed and passes the U-turn at a low velocity. When it finishes turning around, it speeds up to a normal level. As for entering a roundabout, the vehicle always reduces the speed to adapt to different angles of the roads.

\begin{figure}
  \centering
  \subfigure[]{
    \includegraphics[width=2.0in]{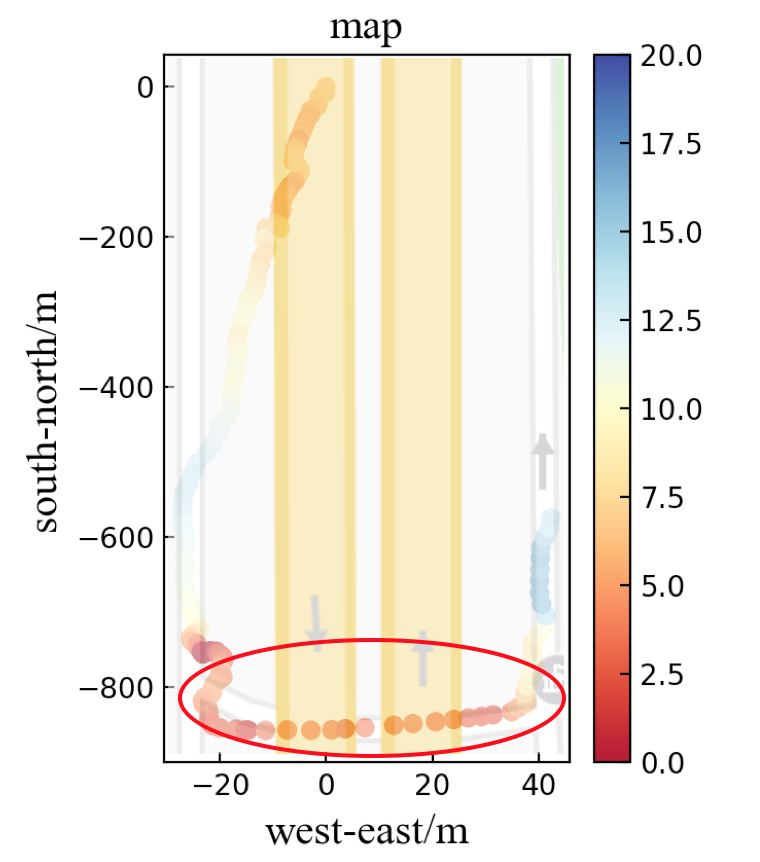}
  }

  \subfigure[]{
    \includegraphics[width=2.5in]{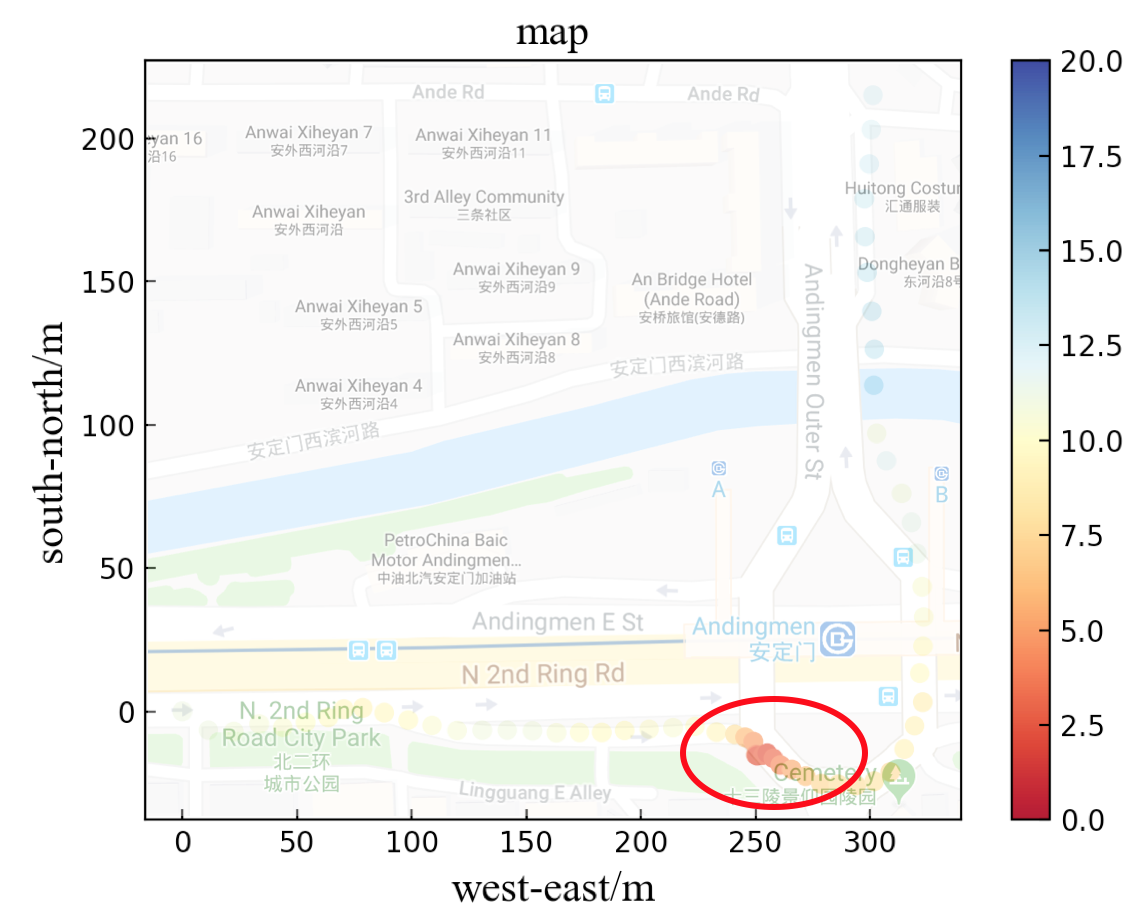}
  }
  \caption{Illustration of the variation of speed and position\protect\\(a). turning around (b). entering a roundabout}
\end{figure}

A KWS system is built to get the results. The system uses DNN model which is trained to recognize keywords like “\emph{HEY, ATOM}”. The baseline result is obtained by testing the single-source system. After that, multi-source fusion scheme is adopted to get the result for comparison.

The sensitivity value of the system can be adjusted from $0$ to $1$. The larger the value is, the more sensitive the KWS system is and the system is more likely to detect keywords and be activated. The acceptable range of sensitivity is $0.495-0.580$. When the sensitivity value is too large, the response rate to negative cases increases sharply. When the sensitivity value is too small, the response rate to positive cases decreases sharply.

\begin{table}[htbp]
\caption{Sensitivity Value for Test}
\begin{center}
\begin{tabular}{|c|c|}
\hline
\textbf{No.}&{\textbf{$sen$}} \\
\hline
1& 0.495 \\
\hline
2& 0.50 \\
\hline
3& 0.52 \\
\hline
4& 0.55 \\
\hline
5& 0.57 \\
\hline
6& 0.58 \\
\hline
\end{tabular}
\label{tab1}
\end{center}
\end{table}

6 sensitivity values are chosen for test, which are listed in Table I. For each value, experiments on KWS system are conducted and results are shown in Fig.6. It shows a relation between sensitivity value and precision, recall rate. With the increase of sensitivity, the precision rate shows a downward trend, and the recall rate shows an upward trend. The precision rate changes from $1$ to $0.8$ while the recall rate changes from $0.3$ to $0.6$.

\begin{figure}
\begin{center}
\includegraphics[width=3in]{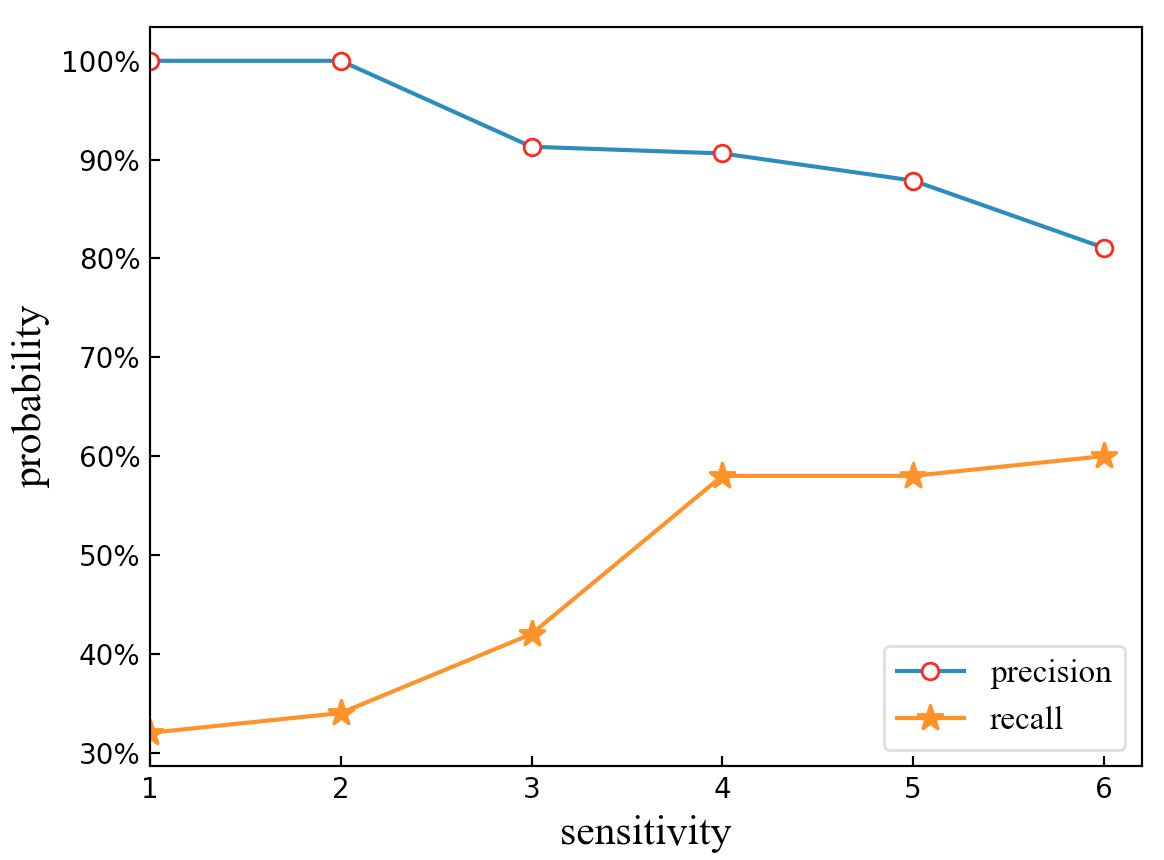}
\caption{The relation between sensitivity value and precision, recall rate} \label{fig6}
\end{center}
\end{figure}

Relatively speaking, most KWS systems have a high precision rate, which means that they are not easy to be activated by words other than keywords. However, recall rate is not always acceptable. When people utter the keyword, system cannot always succeed in activating the system. Therefore, improving the recall rate is usually the main goal of improving the KWS system.

For multi-source fusion scheme, the system can automatically adjust the sensitivity value according to the vehicle information. When the vehicle is moving in a straight line or there is no significant fluctuation in speed, the sensitivity of system is set to be a lower value which is denoted as $sen_1$. When the direction of the vehicle changes and the speed changes significantly, the sensitivity of system is set to be a higher value which is denoted as $sen_2$. 15 sets of double sensitivity values are listed and they are numbered from $1$ to $15$ as is shown in Table II.

\begin{table}[htbp]
\caption{Double Sensitivity Combination}
\begin{center}
\begin{tabular}{|c|c|c|}
\hline
\textbf{No.}&{\textbf{$sen_1$}}&{\textbf{$sen_2$}} \\
\hline
1& 0.495& 0.50 \\
\hline
2& 0.495& 0.52 \\
\hline
3& 0.495& 0.55 \\
\hline
4& 0.495& 0.57 \\
\hline
5& 0.495& 0.58 \\
\hline
6& 0.50& 0.52 \\
\hline
7& 0.50& 0.55 \\
\hline
8& 0.50& 0.57 \\
\hline
9& 0.50& 0.58 \\
\hline
10& 0.52& 0.55 \\
\hline
11& 0.52& 0.57 \\
\hline
12& 0.52& 0.58 \\
\hline
13& 0.55& 0.57 \\
\hline
14& 0.55& 0.58 \\
\hline
15& 0.57& 0.58 \\
\hline
\end{tabular}
\label{tab1}
\end{center}
\end{table}

Each group in Table II is tested and its precision and recall rate are recorded and presented in Fig.7, where the horizontal axis is the sequence number of double sensitivity combination and the vertical axis is the precision and recall rate. After using multi-source fusion scheme, the precision rate changes from $1$ to $0.88$ while the recall rate still changes from $0.3$ to $0.6$. However, it is clear to see that from the 4th group, the recall rate stabilizes at a relatively high level and the precision rate keeps above $0.88$. As a result of the improvement of recall rate, the probability that people utter the correct keyword but fail to activate the system is significantly reduced.

\begin{figure}
\begin{center}
\includegraphics[width=3in]{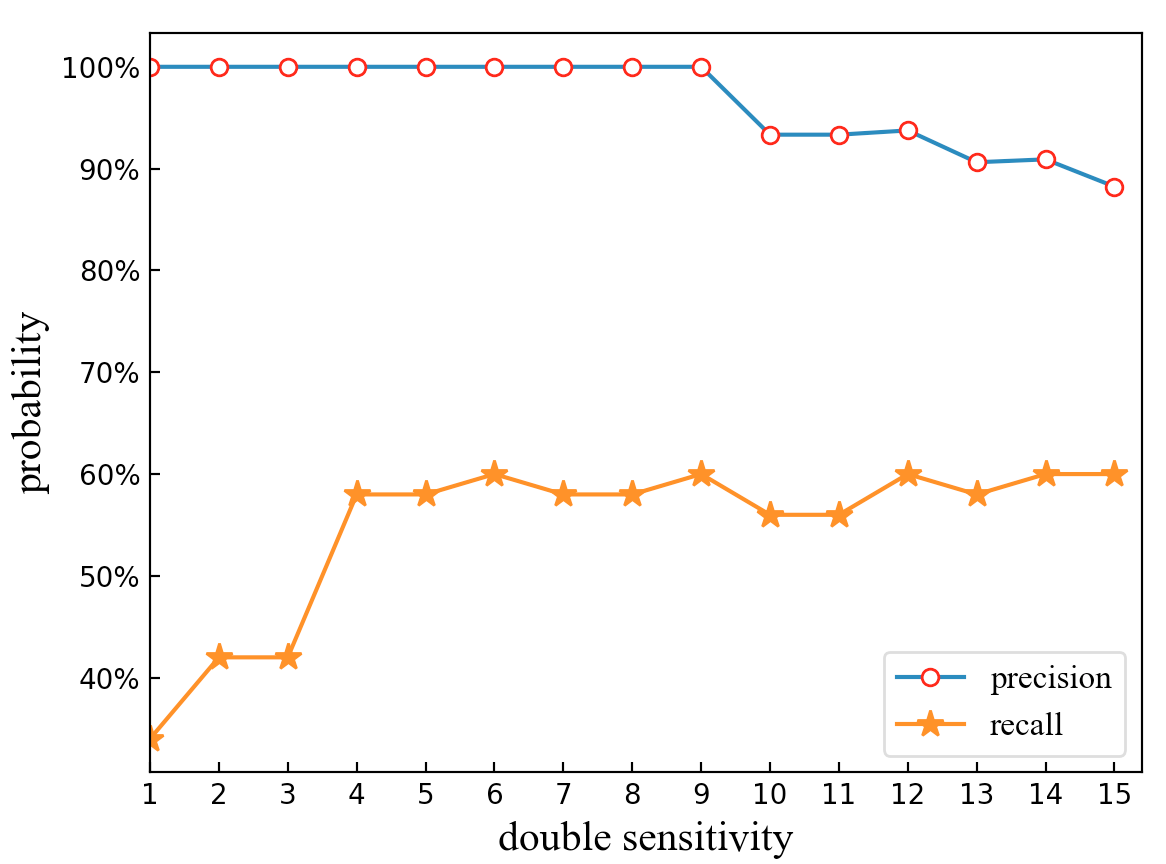}
\caption{The precision and recall rate in different double sensitivity combination} \label{fig7}
\end{center}
\end{figure}

Table III indicates the different mean value and mean square error of precision rate (PR) and recall rate (RR) between single-source and multi-source fusion system. The KWS system with multi-source fusion achieves a $5.3\%$ improvement in mean PR and a $13.0\%$ improvement in mean RR. As is shown in Table III, there is a large gap in mean square error of PR and RR between two systems. A $63.6\%$ reduction in mean square error of PR and $54.4\%$ reduction in mean square error of RR are achieved compared to the single-source system, which increases the availability and reliability of KWS system.

\begin{table}[htbp]
\caption{The Performance of Single-source and Multi-source System}
\begin{center}
\begin{tabular}{|c|c|c|c|c|}
\hline
\textbf{ }&\multicolumn{2}{|c|}{\textbf{Mean Value}}&\multicolumn{2}{|c|}{\textbf{Mean Square Error}} \\
\cline{2-5}
\textbf{ } & \textbf{\textit{precision rate}}& \textbf{\textit{recall rate}}& \textbf{\textit{precision rate}}& \textbf{\textit{recall rate}} \\
\hline
\textbf{sigle-source}& 91.81\%& 47.33\%& 0.00533& 0.01659 \\
\hline
\textbf{multi-source}& 96.68\%& 53.47\%& 0.00194& 0.00757 \\
\hline
\end{tabular}
\label{tab1}
\end{center}
\end{table}

The precision and recall rate of the KWS system can be significantly improved by setting the appropriate double sensitivity value. KWS system with multi-source fusion can choose appropriate parameters according to current state, which enhances the performance of the system.

\section{Conclusions}

In this paper, DNN has been used to detect keyword from a continuous stream of audio and get the precision and recall rate. Based on this, an in-vehicle KWS system with multi-source fusion is proposed for vehicle applications. Vehicle information is used as a complement to the audio information for KWS system. KWS system with multi-source fusion can adjust the sensitivity value for DNN-based single-source KWS system and the experimental results show that automatically adjusting sensitivity value is feasible and optimize the system effectively to achieve better performance in both precision rate and recall rate.

\section*{Acknowledgement}

This work was supported by the National Natural Science Foundation of China (NSFC) under Grant Number 61671089.


\end{document}